\documentclass{article}

     \usepackage[final]{neurips_2019}

\usepackage[utf8]{inputenc} 
\usepackage[T1]{fontenc}    
\usepackage{hyperref}       
\usepackage{url}            
\usepackage{booktabs}       
\usepackage{amsfonts}       
\usepackage{nicefrac}       
\usepackage{microtype}      
\usepackage{natbib}
\usepackage{lipsum}
\usepackage{siunitx}
\usepackage{amsmath}
\usepackage{graphicx}
\usepackage[ruled,linesnumbered]{algorithm2e}
\usepackage{verbatim}

\title{Noisy and Incomplete Boolean Matrix Factorization via Expectation Maximization}

\author{%
 Lifan Liang \\
  Department of Biomedical Informatics\\
  University of Pittsburgh\\
  Pittsburgh, PA 15260 \\
  \texttt{lil115@pitt.edu} \\
  \And
  Lu, Songjian\thanks{} \\
  Department of Biomedical Informatics\\
  University of Pittsburgh\\
  Pittsburgh, PA 15260\\
  \texttt{songjian@pitt.edu}
}

\begin{document}

\maketitle

\begin{abstract}
  Probabilistic approach to Boolean matrix factorization can provide solutions robust against noise and missing values with linear computational complexity. However, the assumption about latent factors can be problematic in real world applications. This study proposed a new probabilistic algorithm free of assumptions of latent factors, while retaining the advantages of previous algorithms. Real data experiment showed that our algorithm was favourably compared with current state-of-the-art probabilistic algorithms.
\end{abstract}

\section{Introduction}
\label{sec:intro}
The problem of Boolean matrix factorization (BMF) is to identify two binary matrices, U and Z, with rank L such that every element in the binary matrix, X, is an OR mixture of AND product:
\begin{equation}
\label{eqn: def}
    Z_{ij} = \lor_{l\leq L} (U_{il} \land Z_{jl})
\end{equation}
where $\lor$ is the OR operator and $\land$ is the AND operator. BMF has found wide application in the area of data mining, including ratings prediction (\cite{ravanbakhsh2016boolean}), boolean databases (\cite{geerts2004tiling}), gene expression bi-clustering (\cite{Zhang2010Binary}), and role mining (\cite{Vaidya2007The}, \cite{Lu2008Optimal}). In this study, we also showed that BMF can be applied to breast cancer subtype classification and three-dimensional segmentation of hippocampal region in mouse brains with solely gene expression profiles.

Although BMF is a NP-hard problem, many efficient approximate algorithms, such as ASSO, (\cite{Miettinen2008The}) have been developed. However, these algorithms is not able to deal with distorted or missing values effectively, which are common issues in real data. Algorithms developed in recent years aimed to handle such issues explicitly. Among these efforts, the probabilistic approach (\cite{neumann2018bipartite}, \cite{ravanbakhsh2016boolean}, \cite{rukat2017bayesian}) is particularly promising. By estimating the uncertainty in probabilistic models, this approach is robust against distorted or missing values.

However, to achieve such robustness and efficiency, these algorithms has made strong assumptions about the shape of the latent factors. Since prior knowledge about the latent factors' shape is usually not available in real data analysis, it is unrealistic to make assumptions over them. From the experience of conducting singular value decomposition, we know it is unlikely that the dominant factors have the same singular values. Although the Boolean rank of real-world data has not been investigated thoroughly, strong assumptions of latent factors' shape probably have imposed bias in the real data analysis.

\begin{figure}
  \centering
  \includegraphics[width=4.5cm, height=4cm]{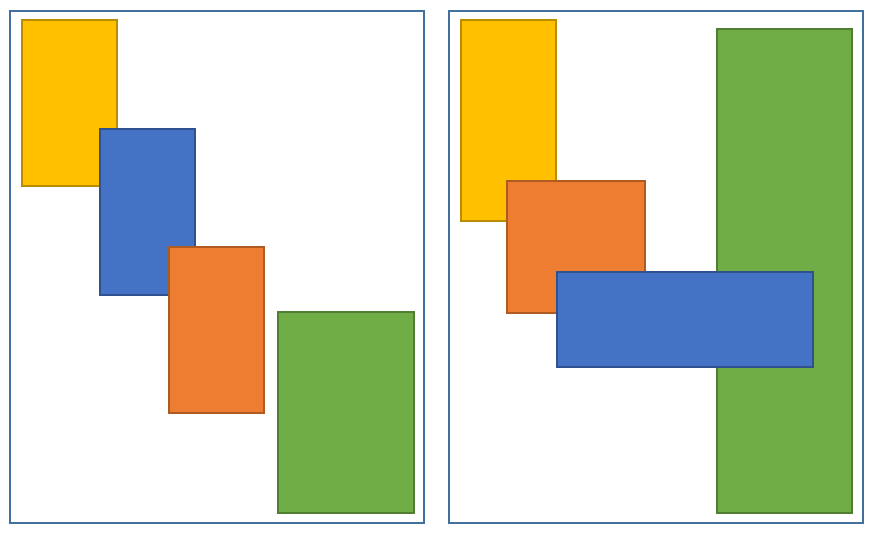}
  \caption{Figure on the left showed a scenario where latent factors comply with uniform Bernoulli prior; figure on the right showed a more realistic scenario where latent factors have various shapes.}
  \label{fig:ass_ill}
\end{figure}

In this study, we presented a new algorithm free of assumptions about Boolean factors while retaining the advantages of previous algorithms. As illustrated in Fig. \ref{fig:ass_ill}, our algorithm aimed to identify Boolean factors accurately in a more realistic scenario where latent Boolean factors can take any shape.

As described in Section \ref{sec: methods}, our approach consists of three novel ideas: (1) allow the latent factors to vary by relaxing the parameters to be continuous values within $[0,1]$ sampled from Beta distribution; (2) reparameterize the parameters from $[0,1]$ to $(-\infty, +\infty)$, thus making simple gradient ascent feasible; (3) uniform noise is directly modeled and jointly estimated with latent factors in an EM algorithm. 

In Section \ref{sec:results}, our algorithm was compared with LoM and message passing. Synthetic experiments showed that our algorithm outperformed both of them when latent factors' sizes varied considerably with each other and observation is not overwhelmed by noise. Real data experiment also indicated that our algorithm has extracted more information with the same number of latent factors.

\section{Related Works}
The BMF problem is closely connected, if not identical, to many other computational problems including dense bipartite subgraph extraction (\cite{lim2015convex},\cite{neumann2018bipartite}), the tiling problem (\cite{geerts2004tiling}), and the binary independent component analysis (\cite{nguyen2011binary}). Various heuristics has been adopted to develop efficient approximates, including the discrete basis problem (\cite{Miettinen2008The}), linear programming (\cite{Lu2008Optimal}), formal concept analysis (\cite{Nenova2011An}, \cite{belohlavek2015below}) and minimum description length (\cite{makhalova2019below}, \cite{Miettinen2011Model}). However, most algorithms cannot handle both noise and missing values. For example, \cite{neumann2018bipartite} presented a simple two-step algorithm that can identify tiny clusters on the right side even under destructive noise level. But it assumed the cluster size on the left are large while those on the right are small. Moreover, it cannot handle missing values.

Thus, our scope narrows down to two algorithms most similar with ours. One is message passing. By modeling the problem as Markov network, \cite{ravanbakhsh2016boolean} has achieved the best performance in noisy matrix factorization and noisy matrix completion respectively. The other is Logical Factorization machine (LoM). LoM (\cite{rukat2017bayesian}) has further improved the performance under high noise level with Bayesian sampling technique. Both algorithms imposed uniform Bernoulli priors over the latent factors.

The major difference between these algorithms and ours, as stated in Section \ref{sec:intro}, is that our algorithm did not assume prior knowledge about the Boolean factors. In addition, to the best of the authors' knowledge, our algorithm is the first to apply gradient ascent to solve Boolean factors on both sides. This enables the algorithm to scale up to huge Boolean matrices with effective implementation.

\section{Model and Implementation}
\label{sec: methods}
\subsection{Problem formulation}
The conventional formulation was described in Eq. \ref{eqn: def}. In this study, a different formulation was adopted. We assume that each element of X, $X_{ij}$, is sampled from a different Bernoulli distribution. Similarly, every element in the latent factors is sampled from different Bernoulli distributions. The generative process of X can be described as follows:
\begin{equation}
    U_{nl} \sim Bernoulli(\mu_{ml})
\end{equation}
\begin{equation}
    Z_{ml} \sim Bernoulli(\zeta_{ml})
\end{equation}
\begin{equation}
    P_{nm} = 1 - P(X_{nm}=0) = 1 - \prod_{l=1}^{L}{(1 - \mu_{nl} * \zeta_{ml})}
\end{equation}
\begin{equation}
    X_{nm} \sim Bernoulli(P_{nm})
\end{equation}
\\
where $\mu$ is a $N \times L$ matrix with values in $[0,1]$, $\zeta$ is a $M \times L$ matrix with values in $[0,1]$. Clearly, by forcing $\mu$ and $\zeta$ to be binary, our formulation will be identical to previous Bayesian approaches. Thus our formulation is a generalized version of previous ones. With this approach, our goal for Boolean matrix factorization is to estimate the parameter $\mu$ and $\zeta$ instead of their samples U and Z.

\subsection{Maximum likelihood estimation}
We estimate $\mu$ and $\zeta$ by maximizing the log likelihood of X, which is:
\begin{equation}
    LL(\mu, \zeta; X) = \sum_{n\leq N,m\leq M} {\big[X_{nm}\log P_{nm} + (1-X_{nm})\log (1-P_{nm})\big]}
\end{equation}
Conventional gradient descent is not application because $\mu$ and $\zeta$ need to be within the interval $[0,1]$. Thus, we reparameterize $\mu$ and $\zeta$ as $\sigma (A)$ and $\sigma (B)$ elementwise:
\begin{equation}
    \mu_{nl} = \frac{1}{1+e^{-A_{nl}}}
\end{equation}
\begin{equation}
    \zeta_{nl} = \frac{1}{1+e^{-B_{ml}}}
\end{equation}

With reparameterization, it becomes a problem of unconstrained nonlinear programming. A simple gradient ascent algorithm is sufficient to jointly optimize the estimators of $A$ and $B$. The partial likelihood gradients regarding A and B are:
\begin{equation}
    \frac{\partial LL}{\partial A_{il}} = 
    \sum_{j\leq m} 
    {\big[ \frac{\mu_{il} \zeta{jl}}{1-\mu_{il} \zeta_{jl}} (1-\mu_{il})(1-\frac{X_{ij}}{P_{ij}}) \big]}
\end{equation}
\begin{equation}
    \frac{\partial LL}{\partial B_{il}} = 
    \sum_{j\leq n} 
    {\big[ \frac{\mu_{jl} \zeta{il}}{1-\mu_{jl} \zeta_{il}} (1-\zeta_{il})(1-\frac{X_{ij}}{P_{ij}}) \big]}
\end{equation}

\subsection{Noise estimation}
We further introduced a parameter, $\epsilon$, to explicitly model the probability that elements in X is contaminated by noise (flipped from 1 to 0 or vice versa). In this scenario, the observed data, $X^*$, is generated as:
\begin{equation}
\label{eqn: noise1}
    C_{ij} \sim Bernoulli(\epsilon)
\end{equation}
\begin{equation}
\label{eqn: noise2}
    X^*_{ij}=
    \begin{cases}
      1-X_{ij}, & \text{if}\ C_{ij}=1 \\
      X_{ij}, & \text{otherwise}
    \end{cases}
\end{equation}
where $C_{ij}$ is a N $\times$ M binary matrix with every element as a i.i.d sample from a Bernoulli distribution parameterized by a scalar $\epsilon$.
To reflect the addition of noise in the model, we need to add one step in the generative process:
\begin{equation}
    P^* = (1-\epsilon)P + \epsilon(1-P)
\end{equation}
The noisy observation, $X^*$, is sampled from $P^*$ instead of $P$:
\begin{equation}
    X^*_{nm} \sim Bernoulli(P^*)
\end{equation}
Thus, the model likelihood becomes:
\begin{equation}
\label{eqn: noise_ml}
    LL(\mu, \zeta, \epsilon; X^*) = \sum_{n\leq N, m\leq M} {\big[X^*_{nm}\log P^*_{nm} + (1-X^*{nm})\log (1-P^*_{nm})\big]}
\end{equation}

To optimize $\mu$, $\zeta$ and $\epsilon$ in Eq. \ref{eqn: noise_ml}, we applied the expectation maximization algorithm. In M step, $\mu$ and $\zeta$ are estimated with the same approach as described in Section 2.2. The difference is the presence of a fixed $\epsilon$, leading to a different equation for likelihood gradients:
\begin{equation}
    \frac{\partial LL}{\partial A_{il}} = \sum_{j\leq m} \Big[(1-\mu_{il})(1-P_{ij})(1-2\epsilon)
    \frac{\mu_{ij}\zeta_{ij}}{1-\mu_{ij}\zeta_{ij}} \frac{P^*_{ij}-X'_{ij}}{(1-P^*_{ij})P^*_{ij}}\Big]
\end{equation}
\begin{equation}
    \frac{\partial LL}{\partial B_{il}} = \sum_{j\leq m} \Big[(1-\zeta_{il})(1-P_{ij})(1-2\epsilon)
    \frac{\mu_{ij}\zeta_{ij}}{1-\mu_{ij}\zeta_{ij}} \frac{P^*_{ij}-X'_{ij}}{(1-P^*_{ij})P^*_{ij}}\Big]
\end{equation}
In E step, based on the modified generative process described in Eq. \ref{eqn: noise1} and Eq. \ref{eqn: noise2}, the expected value of $\epsilon$ is equivalent to the difference between observation, $X^*$, and the reconstructed data without noise:
\begin{equation}
    \epsilon = \frac{|C|}{NM} = \frac{|\Hat X - X^*|}{NM}
\end{equation}
where $|C|$ is the absolute sum of all the elements in $C$. $\Hat{X}$ is reconstructed by the model as:
\begin{equation}
    \Hat{X}=
    \begin{cases}
      1, & \text{if}\ P^*\geq 0.5 \\
      0, & \text{otherwise}
    \end{cases}
\end{equation}

Note that this is an approximate estimate, the exact estimate should be the average difference between $X^*$ and $P^*$. The exact estimate require M-step to reach a much stricter convergence. During synthetic experiments, the performance of approximate estimate is not significantly different from the exact one. Thus the approximate estimate of $\epsilon$ was adopted.

\subsection{MAP Estimation as Regularization}
We further impose prior distribution on $\mu$ and $\zeta$:
\begin{equation}
    \mu_{ml} \sim Beta(\alpha, \beta)
\end{equation}
\begin{equation}
    \zeta_{nl} \sim Beta(\alpha, \beta)
\end{equation}
In practice, $\mu$ and $\zeta$ can comply with different Beta distributions. For the convenience of notation, we simply assume they have a common prior distribution.

Thus $\mu$ and $\zeta$ are estimated based on Maximum a Posteriori (MAP) estimator. The posterior probability function of $\mu$ and $\zeta$ is:
\begin{equation}
    Pr(X|\mu, \zeta, \epsilon) = LL 
    + (\alpha-1) \Big[\sum_{m,l}^{M,L}\log{\mu_{ml}}+\sum_{n,l}^{N,L}\log{\zeta_{nl}}\Big]
    + (\beta-1) \Big[\sum_{m,l}^{M,L}\log{(1-\mu_{ml})}+\sum_{n,l}^{N,L}\log{(1-\zeta_{nl})}\Big]
\end{equation}
where LL is described in Eq. \ref{eqn: noise_ml}. We applied gradient ascent to the objective function. The partial gradient for $Pr(X|\mu, \zeta, \epsilon)$ is:
\begin{equation}
    \frac{\partial Pr(X|\mu, \zeta, \epsilon)}{\partial A_{n,l}} = \frac{\partial LL}{\partial A_{il}}
    + \big(\frac{\alpha-1}{\mu_{n,l}} - \frac{\beta-1}{1-\mu_{n,l}}\big)(1-\mu_{n,l})\mu_{n,l}
\end{equation}
\begin{equation}
    \frac{\partial Pr(X|\mu, \zeta, \epsilon)}{\partial B_{n,l}} = \frac{\partial LL}{\partial B_{il}}
    + \big(\frac{\alpha-1}{\zeta_{n,l}} - \frac{\beta-1}{1-\zeta_{n,l}}\big)(1-\zeta_{n,l})\zeta_{n,l}
\end{equation}
Clearly, when $\alpha$ and $\beta$ are set to 1, the MAP estimator will be identical to the maximum likelihood estimator. When $\alpha$ and $\beta$ are larger than 1, latent factors will be skewed towards 0.5; when $\alpha$ and $\beta$ are less than 1, latent factors are pushed towards 0 or 1. Alternatively, the entropy of $\mu$ and $\zeta$ can be used as penalty and the objective becomes minimizing KL divergence. However, users can push the sparsity of latent factors by making $\alpha$ and $\beta$ asymmetric, which is not available with entropy.

\subsection{Matrix Completion}
As briefly mentioned in Section 2.1, our approach to matrix completion is simple. During training, parameters are only updated based on the gradients from the observed data points. When convergence is reached, missing data are imputed by the reconstructed data without noise.

\subsection{Implementation}
The pseudo code for the model proposed in this study is shown in Algorithm \ref{alg: em}. In addition to the theoretical aspects illustrated in previous sections, here we illustrated several practical decisions based on the algorithm's performance during synthetic experiments: (1) resilient propagation on full batch was adopted to optimize the estimator of latent factors. This is because of its superior performance in terms of convergence rates and optimum loss when compared to vanilla gradient ascent, SGD, and ADAM; (2) the convergence criteria for M-step is whether the reconstructed data is the same as the previous iteration; (3) the priors, $\alpha$ and $\beta$, are set to 0.95 across all the experiments, slightly pushing parameters towards 0/1; (4) the reparameterized parameters, A and B, were clipped after each update, meaning that all of them are bounded within $[-5,5]$. This is necessary given the setting of priors and the convergence criteria.

As for the computation complexity, the most time-consuming step is computing the partial gradient for each element in the factor. The computational complexity in one iteration is $O(NML)$. The size of latent factors, L, is usually small and fixed. Thus the complexity of our algorithm is still linear to the size of the matrix.

\begin{algorithm}[H]
\label{alg: em}
 \SetKwFunction{Gaussian}{Gaussian}\SetKwFunction{Reconstruct}{Reconstruct}\SetKwFunction{ComputeGradient}{ComputeGradient}\SetKwFunction{RPROP}{RPROP}\SetKwFunction{Diff}{Diff}\SetKwFunction{clip}{clip}
 \SetKwInOut{Input}{Input}\SetKwInOut{Output}{Output}

 \Input{X an $N \times M$ binary matrix; L number of latent factors; $\alpha,\beta$, Beta priors}
 \Output{$\mu, \zeta$, latent factors; $\epsilon$, flip probability}
 \BlankLine
 $\epsilon \leftarrow 0$ \;
 $A^{M\times L} \leftarrow Gaussian(mean=0, std=0.01)$\;
 $B^{N\times L} \leftarrow Gaussian(mean=0, std=0.01)$\;
 \While{$|\epsilon - \epsilon^*| > 1e-3$} {
   $\epsilon \leftarrow \epsilon^*$\\
   \While{True} {
     $X' \leftarrow Reconstruct(A, B)$\;
     $G_A^{M \times L}, G_B^{N \times L} \leftarrow ComputeGradient(X, A, B, \epsilon)$\;
     $A^*, B^* \leftarrow RPROP(A, B, G_A, G_B)$\;
     \lIf{$X' == Reconstruct(A^*, B^*)$}{break}
     $A \leftarrow clip(A^*)$\;
     $B \leftarrow clip(B^*)$\;
   }
   $\epsilon^* \leftarrow Diff(X, Reconstruct(A, B))$\;
 }
 \Return $\sigma(A), \sigma(B), \epsilon$
\caption{Expectation Maximization}
\end{algorithm}

\section{Results}
\label{sec:results}
Our algorithm was compared with message passing (\cite{ravanbakhsh2016boolean}) and LoM (\cite{rukat2017bayesian}). The prior for the two algorithms were estimated using empirical Bayes approach described in \cite{rukat2017bayesian}. During synthetic experiments, we evaluated the three algorithms on two tasks: noisy matrix factorization and noisy matrix completion. In real data experiment, the three algorithms were compared on MovieLens datasets and RNAseq datasets from TCGA \cite{tomczak2015cancer}. Finally, we demonstrated our algorithm's real-world application to a spatial transcriptomics dataset.

\subsection{Synthetic Experiment}
The observed matrices with noise, $X^*$, was synthesized based on a sampling scheme as follows:
\begin{equation}
\label{eqn: unif1}
    \omega_l \sim Uniform(p-0.2, p+0.2)
\end{equation}
\begin{equation}
\label{eqn: unif2}
    \theta_l \sim Uniform(p-0.2, p+0.2)
\end{equation}
\begin{equation}
    U_{nl} \sim Bernoulli(\omega_l)
\end{equation}
\begin{equation}
    Z_{ml} \sim Bernoulli(\theta_l)
\end{equation}
\begin{equation}
    X_{nm} = 1 - \prod_{l\leq L} {(1-U_{nl} * Z_{ml})}
\end{equation}
\begin{equation}
    X^*_{nm} = (1-\epsilon)X_{nm} + \epsilon(1-X_{nm})
\end{equation}
The major difference between our sampling scheme and that in previous literature is the variability of the Bernoulli priors of factors, Eq. \ref{eqn: unif1} \& \ref{eqn: unif2}. $p$ was computed from the matrix density $Pr(X=1)$:
\begin{equation}
    Pr(X=1) = 1 - (1-p^2)^L
\end{equation}{}
\subsubsection{Noisy matrix factorization}
We evaluated the three algorithms on five different noise levels (flip probability): 0.0, 0.1, 0.2, 0.3, 0.4. The sampling scheme was repeated 10 times for each noise level. The performance was measured by the reconstruction error rates, which is comparing the reconstructed matrix with the synthesized matrix without noise:
\begin{equation}
    err = \frac{|\Hat X - X|}{NM}
\end{equation}

\begin{figure}
  \centering
  \includegraphics[width=\linewidth, height=2cm]{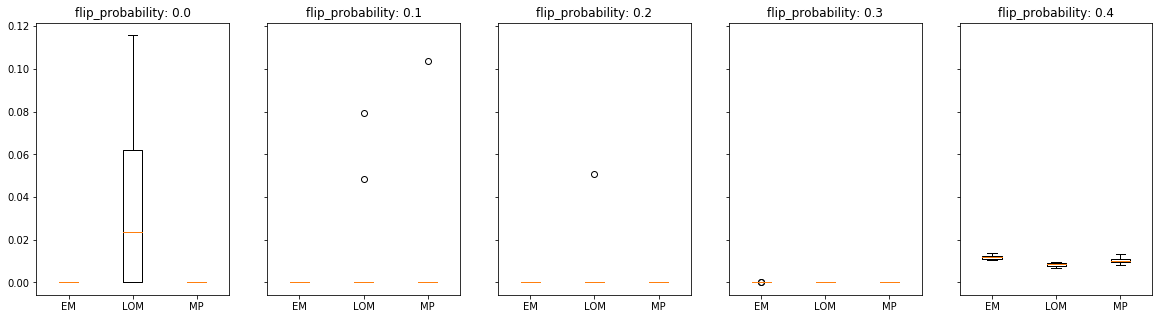}
  \caption{Reconstruction error (12\% max) of synthetic data when Bernoulli priors stayed the same. Synthetic matrices were $1000\times 1000$ with rank 5. EM (Left) is the algorithm proposed in this study; MP (right) is short for message passing; LOM (middle) is the Logical factorization machine.}
  \label{fig:synth_fact_invar}
\end{figure}

\begin{figure}
  \centering
  \includegraphics[width=\linewidth, height=2cm]{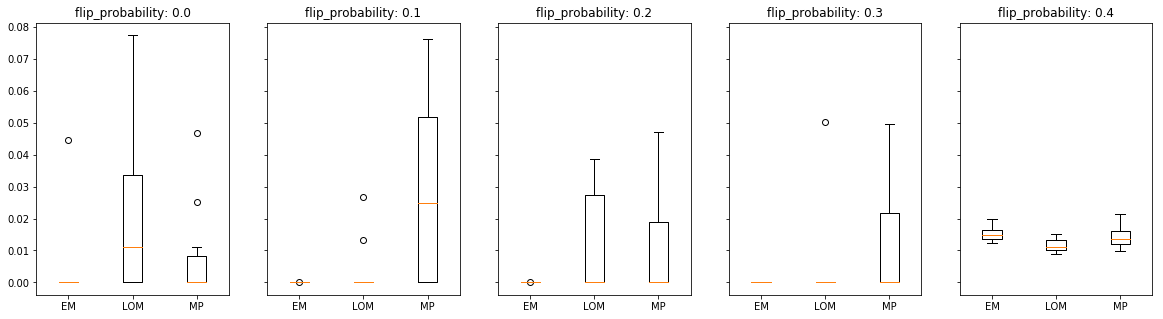}
  \caption{Reconstruction error (12\% max) of synthetic data when Bernoulli priors varied. Synthetic matrices were $1000\times 1000$ with rank 5. EM (Left) is the algorithm proposed in this study; MP (right) is short for message passing; LOM (middle) is the Logical factorization machine.}
  \label{fig:synth_fact_var}
\end{figure}

As shown in Fig.\ref{fig:synth_fact_invar} and Fig.\ref{fig:synth_fact_var}, although EM algorithm is likely to reach a local optimum, the performance of our algorithm is more stable across different noise levels compared with the other probabilistic approaches. Moreover, once Bernoulli priors were allowed to vary, EM has achieved zero error in 9 out of 10 sampled matrices with lower noise levels (flip probability $\leq$ 0.3), while the other two can only perfectly reconstruct the noiseless matrix in 6 to 9 synthetic samples. However, when the flip probability is above 0.3, LoM performed slightly better than message passing and our algorithm. Such comparison result can be observed in matrices with matrix density of 0.3 and 0.7 (shown in supplement Fig 1 \& 2).

When tested against various matrix size and Boolean ranks, the degree of freedom versus sample size ($\frac{(N+M)L}{NM}$) is important for the relative performance of EM. As demonstrated in Supplemental Fig 3 \& 4, when rank was increased from 5 to 10, LoM achieved the best performance across different noise levels. However, when the size of matrix was increased from 1000 to 2500, LoM's performance has a much greater variance than message passing and EM.

\subsubsection{Matrix Completion}
\begin{figure}
  \centering
  \includegraphics[width=\linewidth]{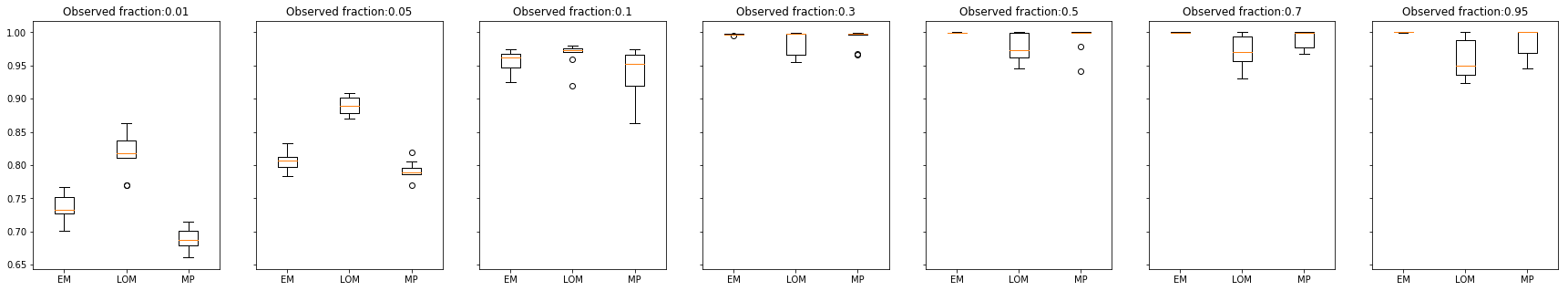}
  \caption{Correctly inferred fraction on synthetic data. EM (left) is the algorithm proposed in this study; MP (right) is short for message passing; LoM (middle) is the Bayesian sampling approach.}
  \label{fig:synth_comp}
\end{figure}

We evaluated the three methods with various observed fraction (i.e. 1\%, 5\%, 10\%, 30\%, 50\%, 70\%, 95\%). The matrices were generated with the same sampling scheme as above. The noise was set at 20\%. The performance was measured by the fraction of correctly inferred values. As shown in Fig. \ref{fig:synth_comp}, LoM accurately inferred 5\% more of the missing data when the fraction of observed data is less than 30\%. However, when the observed fraction has reached above 30\%, the average accuracy of LoM became lower than EM and message passing with greater variance. This is consistent with previous results, indicating that the performance of EM depends heavily on the size of parameters versus the size of observations.

\subsection{Real data experiments}
\subsubsection{Ratings prediction for Movie Lens dataset}
We compared our algorithm with others on the MovieLens datasets \cite{harper2016movielens} with one million ratings and 100,000 ratings respectively. Following previous literature, the ratings were transformed to binary values depending on whether they are above global average. Part of data was randomly chosen to be masked and inferred. The observed fraction ranged from 1\% to 95\%. The experiment was repeated 10 times with each fraction and each algorithm. Each time the observed fraction was resampled. Given in Table.\ref{tab: movielens} was the average performance across the 10 repeated experiments.

\begin{table}
  \caption{Ratings prediction for MovieLens}
  \label{tab: movielens}
  \centering
  \begin{tabular}{lllllll}
    \toprule
        & 1\%   & 5\%   & 10\%  & 20\%  & 50\%  & 95\%   \\
    \midrule
    \textbf{MovieLens-100K}\\
    EM  & \textbf{54.9}  & 58.0  & \textbf{60.3}  & \textbf{62.9}  & \textbf{66.7}  & \textbf{68.2}\\
    MP  & 52.5  & \textbf{58.4}  & 60.2  & 62.6  & 65.0  & 66.4\\
    OrM & 50.8  & 54.0  & 57.8  & 60.9  & 64.2  & 64.7\\
    \textbf{MovieLens-1M}\\
    EM  & \textbf{58.0}  & \textbf{62.5}  & \textbf{64.7}  & \textbf{67.1}  & \textbf{68.8}  & \textbf{69.4}\\
    MP  & 56.2  & 61.9  & 64.1  & 65.7  & 67.5  & 68.4\\
    OrM & 53.2  & 60.9  & 63.2  & 65.0  & 66.4  & 66.8\\
    \bottomrule
  \end{tabular}
\end{table}

\subsubsection{Classification of breast cancer subtypes}
We downloaded gene expression data of breast cancer patients from TCGA (\cite{Tomczak2015The}). The data was dichotomized to encode differential expression. The criteria for differential expression are: (1) absolute log fold change > 0.23; (2) adjusted p value $\leq$ 0.05. Differential expression was encoded as 1, otherwise 0.

From this binary matrix, 15 factors were extracted with our algorithm and others for comparison. Factors about the samples are used as features for tumor subtype classification. Principle component analysis (PCA) was used as baseline. Classification was conducted with Multinomial logistic regression. Performance was evaluated with leave-one-out cross validation. As shown in Table.\ref{tab:accr}, EM algorithm has achieved the highest classification performance among algorithms for Boolean matrix factorization.

We further compared classification accuracy with other Boolean matrix in each tumor subtype. As shown in Table.\ref{tab:sub accr}, all the Boolean matrix factorization methods achieved high accuracy in the subtype of LumA and Basal. It indicates the genes expression data and the subsequent differential expression analysis has provided abundant discriminative information about the two subtypes. However, LoM and Message Passing are not able to discriminate Her2, and Normal-like tumors effectively while EM is somewhat capable of. This result showed that by getting rid of assumptions about factors' sizes, EM is more likely to capture subtle patterns that have greater variance on factor sizes.

\begin{table}
 \caption{Breast cancer subtype classification accuracy}
  \centering
  \begin{tabular}{lll}
    \toprule
    Matrix Factorization    & Accuracy (\%)\\
    \midrule
    EM  & \textbf{81.3}\\
    MP  & 77.7\\
    OrM & 77.8\\
    Baseline    & 50.0\\
    \bottomrule
  \end{tabular}
  \label{tab:accr}
\end{table}

\begin{table}
 \caption{Accuracy for each subtype with 15 factors}
  \centering
  \begin{tabular}{llllll}
    \toprule
    Subtype & Normal & LumA & LumB & Her2 & Basal\\
    \midrule
    \# of Samples &  23 & 417 & 190 & 64 & 140\\
    OrM     & 0.0   & 81.1 & \textbf{74.7} & 48.4 & \textbf{98.5}\\
    MP      & 0.0   & \textbf{91.1} & 53.7 & 53.1 & 94.3\\
    EM      & \textbf{34.8}  & 88.7 & 64.7 & \textbf{65.6} & 96.4\\
    \bottomrule
  \end{tabular}
  \label{tab:sub accr}
\end{table}

\subsubsection{Segmentation of Spatial Transcriptomics}
Spatial transcriptomics data about hippocampal formation in adult mouse brain was downloaded from Allen Brain Atlas (\cite{PMID:17151600}). Our selected region had \~ 5000 voxels. Each voxel contained an expression profile of \~ 20000 genes. Gene expression values were measured with in situ hybridyzation (ISH) technology. As shown in supplement Fig. 5, the number of non-expressed genes was consistent within the same Saggittal section. Thus we believed that most of non-expressed genes are actually missing values and masked them as is. Saggittal sections with less than 3000 expressed genes were removed. Above zero expressions were dichotomized based on individual average of each gene. Clearly, this dataset contained both missing values and noisy measurements, which is suitable to test our algorithm's performance.

Several different sizes of latent factors were attempted, including 2 factors, 5 factors, 10 factors, and 15 factors. As shown in the supplement Fig. 6-9, a range of factor sizes yielded spatially tight cluster without the aids of spatial information.

\bibliography{references}

\end{document}